\documentclass{article}

\usepackage[final]{colm2026_conference}
\usepackage{hyperref}
\usepackage{capt-of}
\usepackage{multicol}
\usepackage{booktabs}
\usepackage{hyphenat}
\usepackage{placeins}
\usepackage{graphicx}
\setkeys{Gin}{draft=false}
\usepackage{microtype}
\usepackage{amsmath}
\usepackage{amssymb}
\usepackage{booktabs,array}
\usepackage{url}
\usepackage{xcolor}
\usepackage[skip=4pt]{caption}
\usepackage{hyperref}
\usepackage{tikz}
\usetikzlibrary{arrows.meta, positioning, calc}
\usepackage{float}
\newcolumntype{C}[1]{>{\centering\arraybackslash}p{#1}}

\definecolor{darkblue}{rgb}{0, 0, 0.5}
\hypersetup{colorlinks=true, citecolor=darkblue, linkcolor=darkblue, urlcolor=darkblue}

\title{Searching for ``Harmful Refusal'': A Psychometric Audit of an AI Safety Benchmark}

\author{Christopher M. Stewart \\
Carnegie Mellon University \\
\texttt{cstewar3@andrew.cmu.edu} \\
\And
Preston Botter \\
Indiana University \\
\texttt{pdbotter@gmail.com} \\
\And
Natalie Sarabosing \\
Carnegie Mellon University \\
\texttt{nsarabos@andrew.cmu.edu} \\
\And
Muye Zhang \\
Google \\
\texttt{andx@google.com} \\
\And
Rachel Phinnemore \\
Google \\
\And
Shalini Ghosh \\
Google \\
\texttt{shalinighosh@google.com} \\
\And
Hong Shen \\
Carnegie Mellon University \\
\texttt{hongs@andrew.cmu.edu} \\
\And
Hoda Heidari \\
Carnegie Mellon University \\
\texttt{hheidari@andrew.cmu.edu} \\
}

\begin{document}

\maketitle

\begin{abstract}
Safety benchmarks typically report one overall score for a suite of datasets, each of which may target one or more safety-related attributes, so models with similar overall scores can have very different attribute profiles. Comparing models is more tractable at the level of individual attributes, yet it is often unclear whether even a single dataset's scores isolate any single attribute. One plausible candidate for such an attribute is \emph{harmful refusal}, a model's tendency to refuse dangerous or policy-violating prompts. We examine whether it constitutes a single, measurable attribute in HELM Safety \citep{helm-safety-2024}. Using a construct validity framework that stipulates that an attribute must exist before a test can measure it \citep{borsboom2004concept}, we start with HELM Safety's four datasets that might plausibly target \emph{harmful refusal}, but find that three are saturated. We subject the remaining dataset, HarmBench \citep{mazeika2024harmbench}, to two psychometric tests to determine if a single attribute like \emph{harmful refusal} could stand behind its score. First, multidimensional item response theory modeling strongly suggests that HarmBench does not measure a singular attribute. Second, a differential item functioning analysis finds items where models from different developers with the same refusal ability score differently. These flags largely disappear under scope-specific matching, a pattern consistent with aggregation effects but not sufficient to rule out domain-specific developer differences.  Zooming out, HarmBench collapses distinct harm behaviors into one score, and the overall HELM safety aggregate further collapses HarmBench and scores from other datasets into a single top-line number. Any safety score that averages over datasets and items can hide saturation and conflate behaviors this way. We argue that a score should earn its single-attribute reading before models are compared with it \citep{salaudeen2025measurement}.
\end{abstract}

\section{Introduction}
AI benchmarks are increasingly treated as measurement instruments, and their scores shape pre-deployment evaluations, procurement guidelines, and model cards, yet those scores are rarely audited for what they measure. The field of psychometrics gives this situation a precise vocabulary. The attribute that a score is meant to measure is called a ``construct'', and a score that no longer tracks its construct has lost ``construct validity''. Under a well-known theorization, a test is valid for measuring a construct if and only if the construct exists and varying it causally produces variation in test outcomes \citep{borsboom2004concept, borsboom2005measuring}. Treating AI benchmarks as measurement instruments requires auditing the inferences that benchmarks support, such as the inference that a higher score means a more capable model. Those inferences are warranted only if a real construct stands behind the score and drives the item responses beneath it \citep{kane2013validity}.

AI safety has typically been described through sets of problematic behaviors \citep{amodei2016concrete, hendrycks2021unsolved}. Benchmark suites operationalize this by grouping datasets on bias, toxicity, privacy leakage, jailbreak robustness, and refusal \citep{yu2026how}. A score built on these separate components sums over behaviors that do not necessarily rise and fall together. Such a score can still rank overall safety, and for some purposes that is enough. However, two models with the same total score can differ sharply on a specific behavior, so a ranking built on that composite value cannot show that one model is safer on any particular attribute \citep{salaudeen2025measurement}. This is a limit on what can be inferred from a composite, and safety decisions often hinge on component-specific claims that one overall score cannot support.

Construct validity becomes a tractable question one level down, at a component dataset that might measure a single construct. There we can ask the prior question that validity depends on, whether the construct exists and whether it alone drives the responses the score aggregates. Among potential AI safety constructs, a model's tendency to decline harmful requests, which we term \emph{harmful refusal}, is a strong candidate. Whether a model declines to comply with harmful requests is a behavioral disposition induced by an identifiable training process, so it names something that could exist and could drive item responses. We put the term in italics here to signal that it is our hypothesized construct and not one that the HELM Safety developers explicitly propose.

HELM Safety \citep{helm-safety-2024} evaluates large language models across five benchmarks and reports a single normalized mean, commonly read as a measurement of model safety. Four of its five datasets are included in an ``LM evaluated harmfulness score'', which scans as similar to our \emph{harmful refusal} concept. These are HarmBench \citep{mazeika2024harmbench}, SimpleSafetyTests \citep{vidgen2023simplesafetytests}, AnthropicRedTeam \citep{ganguli2022redteaming}, and the harmful subset of XSTest \citep{rottger2024xstest}. Sharing the target construct does not guarantee that a dataset measures it, since a dataset can target \emph{harmful refusal} and still fail to discriminate among the pool of models evaluated by HELM Safety. Across the HELM Safety pool, AnthropicRedTeam, SimpleSafetyTests, and the harmful XSTest subset have mean pass rates between 0.92 and 0.94, with almost every model passing almost every item. Of course, a saturated dataset may still measure the construct well for weaker models. However, only HarmBench retains enough variation in this pool to model at the item level, so it alone lets us test whether \emph{harmful refusal} drives the responses.

Per the first condition of Borsboom's construct validity framework, HarmBench can only measure \emph{harmful refusal} if such a construct exists. We probe this question from two sides, using standard psychometric diagnostics \citep{truong2026aims}. The first looks inside the response matrix, that is the table of which models passed which items. A latent dimension is an axis, estimated from the responses, along which models differ. If a single construct drives the responses, one such dimension should account for how models pass and fail. Instead, we find that several latent dimensions predict held-out responses far better than one. The second test incorporates which developer built each model. If \emph{harmful refusal} alone drives the responses, the developer that built a model should not affect its score. Matched on overall refusal ability, models from different families still diverge on a handful of items. That divergence is reduced under scope-specific matching, which is consistent with aggregation effects but does not exclude real domain-specific developer differences. A single latent dimension does not account for HarmBench's responses in this pool, so the aggregate score does not isolate one \emph{harmful refusal} construct. The rest of this paper develops both tests in full and draws out what they mean for reading safety benchmarks.

\section{Construct validity tests}
\label{sec:cvtests}

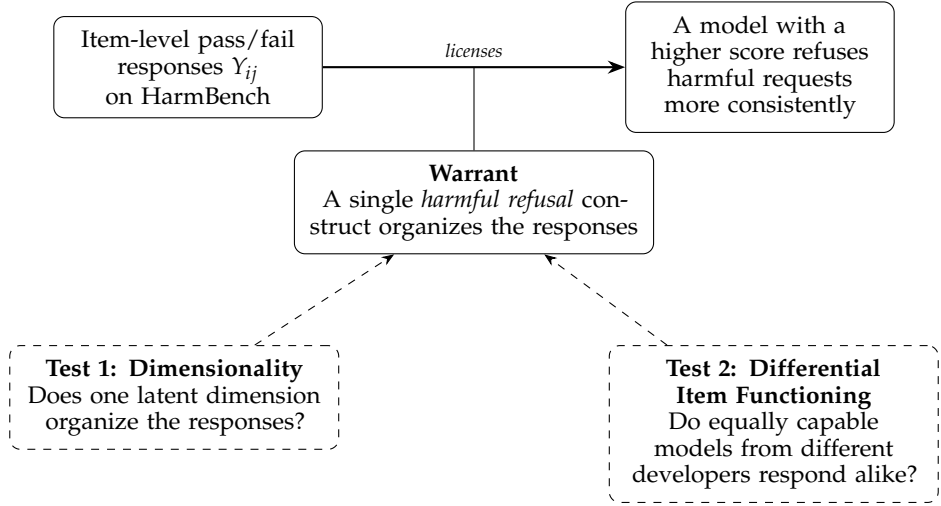
\begin{figure}[htbp]
\centering
\begin{tikzpicture}[
  >=Stealth,
  font=\small,
  box/.style={draw, rounded corners, align=center, inner sep=5pt, text width=32mm},
  wbox/.style={draw, rounded corners, align=center, inner sep=5pt, text width=44mm},
  tbox/.style={draw, dashed, rounded corners, align=center, inner sep=5pt,
               text width=40mm, font=\footnotesize},
  lbl/.style={font=\scriptsize\itshape}
]
  \node[box] (data) {Item-level pass/fail responses $Y_{ij}$ on HarmBench};
  \node[box, right=40mm of data] (claim)
        {A model with a higher score refuses harmful requests more consistently};
  \draw[->, thick] (data) -- node[above, lbl] {licenses} (claim);

  \coordinate (mid) at ($(data)!0.5!(claim)$);
  \node[wbox, below=11mm of mid] (warrant)
        {\textbf{Warrant}\\A single \emph{harmful refusal} construct organizes the responses};
  \draw[-] (mid) -- (warrant);

  \node[tbox, below left=12mm and -6mm of warrant] (t1)
        {\textbf{Test 1: Dimensionality}\\Does one latent dimension organize the responses?};
  \node[tbox, below right=12mm and -6mm of warrant] (t2)
        {\textbf{Test 2: Differential Item Functioning}\\Do equally capable models from different developers respond alike?};
  \draw[->, dashed] (t1) -- (warrant);
  \draw[->, dashed] (t2) -- (warrant);
\end{tikzpicture}
\caption{The HarmBench score supports a claim about a model only through a warrant. The warrant holds that one \emph{harmful refusal} construct organizes the item responses. Our two construct validity tests probe that warrant, one from inside the response matrix and one from outside it.}
\label{fig:warrant}
\end{figure}

Alongside token counts and annotator diagnostics, HELM Safety reports one HarmBench score that represents a model's performance. That score averages the item-level judgments into a single number. A \href{https://crfm.stanford.edu/2024/11/08/helm-safety.html}{blog post} accompanying the leaderboard describes HarmBench as a ``measure of the efficacy of jailbreaking techniques'' that spans several harm domains. It includes a brief description of the dataset and a link to the relevant publication. Following the suggestions of the HELM Safety blog, we read their ``LM evaluated harmfulness score'' as being a similar, if not identical, construct to our \emph{harmful refusal}.

Connecting a HarmBench score to a broader concept like \emph{harmful refusal} requires reading meaning into the raw data. On its own, the score is only a tally of which prompts a model passed and which it failed. Reading it as a measure treats that tally as a sign that the model has a tendency to refuse harmful requests. Measurement researchers break an inference like this into the following three parts \citep{truong2026aims}. The data are the item-level pass and fail responses. The claim is that a model scoring higher refuses harmful requests more consistently. The warrant is what connects the data to the claim. Because the claim is about one construct, it only holds if one construct drives the responses, and here that is \emph{harmful refusal}. This is the fragile part. If several constructs drive the responses instead, the score still exists, but it now measures a blend of them rather than \emph{harmful refusal} alone.

We test the warrant by checking two consequences that should follow if it holds. The first consequence is visible inside the response matrix. In this context, the construct is the theoretical attribute we intend to measure, while the latent dimension is the unobserved statistical axis we estimate from the data to represent it. If one construct drives the responses, one latent dimension should account for the bulk of how models pass and fail. If several latent dimensions are needed instead, the items are answering to more than one construct, and a single score adds those constructs together rather than measuring exactly one. We use dimensionality tests to investigate this question. The second consequence uses metadata completely absent from the response matrix: the identity of the AI model's developer. If \emph{harmful refusal} alone drives the responses, then two models equally matched on overall refusal ability should answer any given item the same way. Their developer should not matter. If equally able models from different developers pass an item at different rates, the response depends on something other than \emph{harmful refusal}. Psychometrics terms this ``construct-irrelevant variance'', and detecting it is the goal of ``differential item functioning'' (DIF). We test this with the DIF analysis below. If one or both consequences do not obtain, the claim that the score measures \emph{harmful refusal} loses its warrant, even though narrower claims about particular behaviors tested in HarmBench items can still hold \citep{salaudeen2025measurement}.

Both tests require modeling item behavior, not just inspecting the response matrix or reading HarmBench's category labels. When models exhibit divergent response patterns across different sets of items, there are two competing explanations: either the items measure entirely different constructs, or they measure the same construct but at different levels of difficulty. HarmBench's behavior categories are useful hypotheses about item content and response process, but they do not by themselves show what underlying abilities explain the observed pattern of passes and failures. Item Response Theory (IRT) separates these possibilities by estimating each item's difficulty and discrimination, then asking whether structured differences among items remain once those properties are accounted for. The same logic makes the second test possible. Comparing developers is meaningful only after their models are mathematically matched on a baseline refusal ability, and IRT provides the exact framework needed to estimate that underlying trait.

\section{Data and methods}
\label{sec:methods}
 
\subsection{Data and scope}
We use the most recent HELM Safety release at the time of writing (v1.17.0, accessed on 21 June 2026), which reports per-item judge scores for every evaluated model. Each item receives a continuous safety score averaged from two LLM judges (\texttt{gpt-4o-2024-05-13} and \texttt{llama-3.1-405b-instruct-turbo}) evaluating on a five-point rubric. We binarize strictly at 1.0, treating only a unanimous perfect score as a pass. This yields a conservative dichotomous response matrix. Under this binarization, HarmBench loses 2 items of 400. Whereas the saturated \emph{harmful refusal} datasets have pass rates between 0.92 and 0.94, HarmBench has a pass rate of 0.67.
 
HarmBench's prompts attempt to elicit harmful behavior, and each item is scored on whether the model complied or refused \citep{mazeika2024harmbench}. HarmBench supplies two item taxonomies. First, three functional categories describe the response process. Standard behaviors are direct harmful requests (199 items), contextual behaviors embed the request in a longer scenario (99 items), and copyright behaviors ask the model to reproduce protected text (100 items). Second, seven semantic categories describe content, including cybercrime (67 items), chemical or biological harm (56 items), copyright violations (100 items), misinformation (65 items), harassment (25 items), illegal activities (63 items), and general harm (22 items). These two taxonomies overlap, but not perfectly. Most semantic categories feature both standard and contextual prompts. The exception is the ``copyright'' category, which exists in both taxonomies, so every copyright item shares the same functional and semantic classification.

\subsection{Models and estimation}
\label{sec:models}
Both tests ask whether one latent ability is enough to organize the HarmBench response matrix, or whether several are needed. Multidimensional item response theory (MIRT) can represent either case, a single latent ability or several, while still estimating each item's difficulty and discrimination \citep{birnbaum1968some}. In an exploratory two-parameter logistic (2PL) MIRT model, every item may load on every dimension, and the probability that model $i$ passes item $j$ takes the form

$$
P(Y_{ij} = 1 \mid \theta_i, a_j, b_j) =
\sigma\left(\sum_{k=1}^{d} a_{jk}(\theta_{ik} - b_{jk})\right),
$$
where $\theta_i$ is model $i$'s latent ability vector, $a_j$ is item $j$'s discrimination vector, $b_j$ is its difficulty vector, and $\sigma$ is the logistic function. Exploratory MIRT is analogous to unsupervised learning in that the method receives the item-response matrix but no preassigned item-to-factor labels, and it estimates latent dimensions directly from the full pattern of pass/fail responses. We fit exploratory 2PL models from $d=1$ through $d=10$, but interpret them mainly as a test of whether one dimension is too few.
 
We fit two confirmatory MIRT models. The first is a three-dimensional response-process model in which standard, contextual, and copyright items each load on their own factor. The second is a seven-dimensional HarmBench-domain model in which each item loads on its semantic harm domain. We treat the three-factor response-process model as primary because it is the more parsimonious named structure. The seven-factor HarmBench-domain model is a sensitivity check that asks whether HarmBench's own harm-domain taxonomy changes the conclusion. Both confirmatory models estimate correlations among their named factors, which allows us to ask if the factors are meaningfully distinct.
 
We compare the IRT models by held-out prediction across five repeated 80/20 response-level train/test splits. Within each model--split cell, we fit twenty random initializations and retain the fit with the highest training log-likelihood; held-out responses are never used for restart selection. We report the mean and standard deviation across the five retained split-level fits. For the direct comparison between the strongest unidimensional specification and the confirmatory 3D model, we also summarize within-split variation across the twenty starts as an optimization-stability diagnostic. Held-out log-loss is the primary criterion, with Brier score as a calibration-sensitive secondary check. Both measure how well a model's predicted pass probabilities match the actual outcomes, with lower values better. AIC and BIC are reported as in-sample parsimony checks. The HarmBench matrix contains 81 models and 398 items ($n < p$). We use variational inference in py-irt \citep{lalor2023pyirt} and retain binary 2PL as the primary, simpler response model. A graded-response model preserves score levels but adds item thresholds, each informed by the same small model pool. Their recovery under sparse categories and dependent model lineages requires validation beyond predictive fit \citep[Problems 1 and 3]{botter2026benchmarks}. This motivates a limited scope, not a claim that 2PL avoids these problems. Implementation and seed details are in Appendix~\ref{app:repro}. Model-family checks are in Appendix~\ref{app:estimation}.
 
\paragraph{Dimensionality} We organize the dimensionality analysis in three parts. First, a Bernoulli-null eigenvalue check works on the item-correlation matrix. This matrix captures whether the models that pass one item tend to pass another. It asks whether that matrix is compatible with one common factor. This is a model-light diagnostic rather than an IRT fit. Second, exploratory MIRT fits 1D through 10D 2PL models with no item-to-factor labels. These models ask whether extra latent dimensions predict held-out responses better than a single scale, but they do not by themselves name the constructs. Third, confirmatory MIRT fits the 3D response-process model and the 7D HarmBench-domain model. These theory-constrained models ask whether interpretable, pre-specified factors explain the predictive gains parsimoniously, and their estimated factor correlations show whether the specified factors are practically interchangeable.

\paragraph{Differential item functioning} DIF asks whether an item is calibrated the same way across groups, rather than which developer's models are safer on average. Overall differences in safety do not constitute differential functioning. DIF flags an item-level group difference after conditioning on estimated ability; it does not establish why the difference occurs. The analysis is therefore strictly conditional. If \emph{harmful refusal} is the sole construct an item measures, group identity should not affect the probability of a pass once models are matched on estimated ability. An item where developer identity still shifts this probability exhibits differential functioning.
 
The groups are fixed in advance. We restrict the developer screen to groups with at least eight models, leaving OpenAI and Anthropic with 21 and 11 models. This is a pragmatic inclusion rule, not a validated sample-size requirement. We also compare closed or API-accessed models with open-weight-like models. Our primary DIF screen is Mantel-Haenszel (MH) within three ability bands \citep{hollandwainer1993dif}. MH first groups models by their fitted refusal ability, then compares groups only within the same ability band. For each item, it asks whether models from one group are still more or less likely to pass than similarly able models from the other group. We calibrate flags by simulating 5{,}000 no-DIF response matrices from the fitted IRT model with group labels fixed, then use the maximum simulated statistic as a family-wise cutoff. A ridge-penalized logistic screen uses continuous fitted ability rather than ability bands as a sensitivity check \citep{zumbo1999handbook}. We run both screens under the single score, the three response-process scores, and the seven HarmBench-domain scores.

We retain MH and ridge-logistic DIF as simpler conditional screens rather than extend the analysis to multiple-group IRT with anchor purification. With only 21 OpenAI and 11 Anthropic models, the adequacy of group-specific item estimation, anchor selection, and uncertainty calibration is not established here. These remain open issues for small, dependent LLM groups \citep[Problem 2]{botter2026benchmarks}; the simpler screens are not exempt from them.
 
\section{Results}
\label{sec:results}

All results below are relative to the HELM Safety v1.17.0 model pool. Saturation and the dimensional structure that we report are properties of these datasets evaluated against these models, not claims about the benchmarks in every possible cohort.
 
\paragraph{Dimensionality.} The dimensionality evidence comes from three sources: a Bernoulli-null eigenvalue check, exploratory MIRT, and confirmatory MIRT. First, the model-light item-correlation diagnostic rejects a clean one-factor account. It compares observed eigenvalues with a random-data cutoff from Bernoulli matrices matched to HarmBench's shape and item pass rates. Under one factor plus noise, only the leading eigenvalue should clear the cutoff. However, several later eigenvalues do as well. This is evidence against unidimensionality, not an exact factor count. A supporting plot can be found in Appendix~\ref{app:parallel-analysis}.
 
Second, exploratory MIRT gives the same answer with more nuance (Table~\ref{tab:comparison}). The 1D 2PL reaches a held-out log-loss of 0.470 (SD 0.003). Every exploratory model from 2D through 10D improves sharply, with mean log-loss around 0.28. But the exploratory sequence does not identify a stable best dimensionality, and AIC and BIC grow quickly as parameters accumulate. Exploratory MIRT therefore tells us that one dimension is too weak, but not which constructs the additional dimensions represent.
 
\begin{table}[htbp]
\centering
\scriptsize
\setlength{\tabcolsep}{3pt}
\renewcommand{\arraystretch}{1.05}
\caption{HarmBench model comparison across five repeated 80/20 response-level holdout splits. Log-loss and Brier entries are mean (SD) across splits after selecting one fit per model and split from twenty random starts by training log-likelihood; held-out responses were not used for restart selection. All models are 2PL except the unidimensional 3PL row included as the strongest unidimensional comparator. Exploratory models allow all items to load on all dimensions. Confirmatory 3D fixes standard, contextual, and copyright items to response-process factors. Confirmatory 7D fixes items to HarmBench harm-domain factors. Lower is better in the fit columns. Bold marks the best value in each column. AIC and BIC are means across splits and are rounded.}
\label{tab:comparison}
\begin{tabular}{lrrrrrr}
\toprule
Model & $d$ & Params & Log-loss (SD) & Brier (SD) & AIC & BIC \\
\midrule
Unidimensional (1D 2PL) & 1 & 877 & 0.470 (0.003) & 0.156 (0.001) & 25{,}806 & 32{,}961 \\
Unidimensional (1D 3PL) & 1 & 1{,}275 & 0.322 (0.007) & 0.096 (0.002) & 17{,}600 & 28{,}001 \\
Exploratory 2PL & 2 & 1{,}754 & 0.287 (0.004) & 0.087 (0.001) & 16{,}112 & 30{,}421 \\
Exploratory 2PL & 3 & 2{,}631 & 0.284 (0.006) & 0.086 (0.002) & 17{,}782 & 39{,}245 \\
Exploratory 2PL & 4 & 3{,}508 & 0.285 (0.007) & 0.086 (0.002) & 19{,}490 & 48{,}108 \\
Exploratory 2PL & 5 & 4{,}385 & 0.282 (0.007) & 0.085 (0.002) & 21{,}329 & 57{,}100 \\
Exploratory 2PL & 6 & 5{,}262 & 0.284 (0.007) & 0.085 (0.002) & 23{,}089 & 66{,}015 \\
Exploratory 2PL & 7 & 6{,}139 & 0.285 (0.006) & 0.085 (0.001) & 24{,}866 & 74{,}946 \\
Exploratory 2PL & 8 & 7{,}016 & 0.284 (0.007) & 0.086 (0.001) & 26{,}809 & 84{,}043 \\
Exploratory 2PL & 9 & 7{,}893 & 0.282 (0.005) & 0.085 (0.001) & 28{,}459 & 92{,}848 \\
Exploratory 2PL & 10 & 8{,}770 & 0.284 (0.006) & 0.086 (0.002) & 30{,}300 & 101{,}843 \\
\midrule
Confirmatory 3D & 3 & 1{,}039 & 0.258 (0.006) & 0.078 (0.002) & \textbf{13{,}707} & \textbf{22{,}183} \\
Confirmatory 7D & 7 & 1{,}363 & \textbf{0.255 (0.006)} & \textbf{0.077 (0.002)} & 13{,}961 & 25{,}080 \\
\bottomrule
\end{tabular}
\end{table}

Third, confirmatory MIRT supplies the interpretable alternatives. The HarmBench-domain 7D model is the best predictive model by a small margin, with a held-out log-loss of 0.255 (SD 0.006). The response-process 3D model is nearly identical at 0.258 (SD 0.006), while using fewer fitted quantities and winning AIC and BIC. More importantly for the unidimensional comparison, the strongest unidimensional specification is the 3PL at 0.322 (SD 0.007). Relative to that model, confirmatory 3D reduces log-loss by 0.064 on average (SD 0.003; range 0.060--0.067), and all five splits favor 3D. Brier score likewise falls from 0.096 (SD 0.002) to 0.078 (SD 0.002), a paired reduction of 0.0176 (SD 0.0005). The comparison is also stable across the twenty starts within each split. The mean within-split SD of held-out log-loss is 0.00037 for the unidimensional 3PL and 0.00035 for confirmatory 3D, and the restart ranges do not overlap within any split: every confirmatory 3D restart outperforms every unidimensional 3PL restart. 

Parsimony settles which model is primary. The 3D standard/contextual/copyright structure is preferred because it predicts almost as well as the 7D model while using far fewer fitted quantities. The 7D harm-domain model is retained as a sensitivity check because it preserves HarmBench's content taxonomy and confirms the same construct validity point. In sum, HarmBench responses are not organized by one \emph{harmful refusal} dimension.
 
The confirmatory factor correlations show that the named factors are related but not interchangeable. In the 3D model, standard and contextual factors correlate at 0.785, while copyright correlates with them at 0.451 and 0.603. In the 7D model, non-copyright harm domains are generally more tightly correlated, often between about 0.65 and 0.85, while copyright is lower, ranging from 0.34 to 0.64 against the other domains. The separation is therefore not just a predictive artifact. The interpretable factors preserve meaningfully different patterns of model behavior, especially around copyright reproduction.

\paragraph{Follow-up analysis without copyright.}
The correlation of 0.785 between the standard and contextual factors, compared with their weaker correlations with copyright, raises the possibility that copyright accounts for the dimensional separation, while the remaining items reflect one attribute with different item difficulties. To examine this possibility, we removed
copyright items and compared a unidimensional 2PL with a confirmatory
2D 2PL assigning standard and contextual items to separate factors.
Both models allowed item-specific difficulty and discrimination. We compare against the 1D 2PL rather than the 3PL here because the confirmatory 2D 2PL nests it, so the comparison isolates the effect of the second dimension. We generated five new 80/20 response-level splits for this subset, shared between the two models, and selected among twenty starts per model and split using training log-likelihood. Excluding items with constant training responses left 295--298 items per split. The 2D model achieved lower held-out log-loss in all five splits: 0.278 (SD 0.007), compared with 0.293 (SD 0.009) for 1D. The paired reduction averaged 0.0149 (SD 0.0043).
Brier score also improved in every split, averaging 0.0851 versus
0.0891. Thus, the predictive advantage of separating standard and
contextual responses persists without copyright, even when the
single-dimensional model allows items to differ in difficulty and
discrimination. This supports residual dimensional structure, but
does not by itself establish two distinct substantive constructs.
 
Supplemental model-family checks point the same way. Rasch/1PL variants do not rescue a one-dimensional score, and 3PL variants with lower asymptotes do not improve the preferred 2PL confirmatory models on held-out log-loss. The full robustness table appears in Appendix~\ref{app:estimation}.
 
\paragraph{Differential item functioning.} The DIF results apply the calibration logic from the Methods section item by item. If an item measures only \emph{harmful refusal}, OpenAI and Anthropic models with the same estimated refusal ability should pass it with equal probability. Several HarmBench items fail this check. Under the single score, OpenAI and Anthropic models matched on overall ability still differ on 13 items at the family-wise Mantel-Haenszel cutoff and 17 under the logistic sensitivity check (Table~\ref{tab:dif}).

\begin{table}[htbp]
\centering
\scriptsize
\setlength{\tabcolsep}{4pt}
\renewcommand{\arraystretch}{1.05}
\caption{Family-wise DIF flags after matching models on the relevant ability estimate. Rows include the single score, 3D response-process scopes, and 7D HarmBench-domain scopes. MH is the Mantel-Haenszel screen. Logit is the ridge-logistic sensitivity screen. Cutoffs are the 95th percentile of the maximum statistic across 5{,}000 bootstrap samples simulated with no group differences present.}
\label{tab:dif}
\begin{tabular}{lrrrrr}
\toprule
& & \multicolumn{2}{c}{OpenAI vs.\ Anthropic} & \multicolumn{2}{c}{Closed vs.\ open} \\
\cmidrule(lr){3-4}\cmidrule(lr){5-6}
Scope & Items & MH & Logit & MH & Logit \\
\midrule
Single score & 398 & 13 & 17 & 0 & 0 \\
\midrule
\multicolumn{6}{l}{3D response-process scopes} \\
Standard & 199 & 0 & 2 & 0 & 0 \\
Contextual & 99 & 1 & 0 & 0 & 0 \\
Copyright & 100 & 0 & 0 & 0 & 0 \\
\midrule
\multicolumn{6}{l}{7D HarmBench-domain scopes} \\
Chemical/biological & 56 & 0 & 0 & 0 & 0 \\
Copyright & 100 & 0 & 0 & 0 & 0 \\
Cybercrime/intrusion & 67 & 2 & 4 & 0 & 0 \\
Harassment/bullying & 25 & 0 & 0 & 0 & 0 \\
General harmful content & 22 & 0 & 0 & 0 & 0 \\
Illegal behavior & 63 & 0 & 0 & 0 & 0 \\
Misinformation/disinformation & 65 & 0 & 1 & 0 & 0 \\
\bottomrule
\end{tabular}
\end{table}
 
Once standard, contextual, and copyright items are scored separately in the 3D response-process model, OpenAI-versus-Anthropic flags collapse to 1 under Mantel-Haenszel and 2 under the logistic check. The 7D screens are very similar in substance. Most domains have no family-wise developer DIF after matching. The only notable residual case is cybercrime/intrusion, with 2 Mantel-Haenszel and 4 logistic flags, plus 1 logistic flag in misinformation/disinformation. In sum, broad developer-linked DIF appears under the single score, then mostly disappears once models are matched within a more specific response-process or harm-domain score.

The reduction in flags is consistent with aggregation effects, but does not establish their cause. Scope-specific ability estimates use the same items being tested and can absorb domain-specific developer differences. We therefore interpret the DIF findings as conditional on the matching procedure, not as evidence that genuine developer differences are absent \citep[Problem 2]{botter2026benchmarks}.

The provenance comparison reinforces the point. It shows no family-wise flags under the single score, the 3D scores, or the 7D scores, even though closed/API and open-weight-like models differ widely in mean score, 0.776 against 0.507, because DIF compares groups only after ability matching. The result is not that one model group is globally safer. It is that detected item-level group differences depend on which ability estimate is used for matching. 

\section{Discussion}
\label{sec:discussion}
Our finding is about HarmBench, but the problem it exposes is not. A single safety score sums over behaviors that do not necessarily move together. Two models with the same score can have very different strengths and weaknesses across those behaviors. A developer difference can then come from the way the score lumps behaviors together, not from one developer being safer. Nothing in either test is specific to HarmBench. Both can be run on any benchmark that publishes item-level responses, since dimensionality needs only the response matrix and ability-matched DIF needs only a grouping like model family. The checks are cheap and fast enough to run on any model release. A pointer to the code to run the analyses can be found in the Appendix.

There are two limits to our claims that deserve mention. First, we tested whether a single \emph{harmful refusal} construct exists, not whether it causally produces the scores \citep{borsboom2004concept}. We do not need the second test here. Because the responses are multidimensional in this pool, the single-score interpretation fails regardless of the causal condition, so the multidimensionality settles the validity question for that interpretation. Second, our evidence rejects a single dimension rather than fixing an exact count of dimensions. Three and seven dimensions predict about equally well, so the data do not single out a true count. What matters for our claim is that one dimension is not enough, and we report the three and seven-factor models because they are the most parsimonious structures that capture the separation.

Looking inside the HarmBench dataset, the copyright items provide the sharpest illustration of why the single score misleads. Whether a model reproduces protected text depends mostly on what it memorized during training, not on whether it judges a request harmful and declines. Refusing such prompts can still matter, since reproduction can expose a developer to real harm, but it is a different behavior from declining a dangerous request that is driven by different facts about the model. This is probably why it separates so cleanly from the other two item types. A score that averages the copyright items together with the others is mixing two different drivers, refusal disposition and memorization, into one number. The separation is not a quirk of the data. Rather, it is what the construct predicts once the label is taken apart.

More broadly, this audit is a small instance of the validity work that rigorous AI evaluation requires. Although benchmark scores already shape procurement guidelines, model cards, and deployment decisions, developers and users have invested comparatively little effort in nailing down the construct that produces them. Measurement science treats a reported score as a claim to be checked rather than a fact to be taken at face value. Here we have shown that, for the current pool of models, HarmBench's composite score cannot be read as measuring a single construct like \emph{harmful refusal}. Because HarmBench is the only dataset still separating models in the latest HELM Safety pool, HELM Safety's \emph{harmful refusal} signal rests on a benchmark whose score does not isolate one attribute. The single score does not measure one thing, so the safest use is the narrow one that keeps the behaviors separate. The broader takeaway is that a number offered as a measure of one attribute should earn that reading before it is used to compare models.

\section{Conclusion}
\label{sec:conclusion}

AI safety leaderboards report scores for constructs like \emph{harmful refusal}, a model's ability to decline dangerous or policy-violating prompts. However, it is often unclear exactly what those scores measure. We took \emph{harmful refusal} as the strongest candidate for a single safety construct and asked whether we could find the construct in the HELM Safety dataset. We could not. Three of the four relevant datasets were saturated. To probe for \emph{harmful refusal} in the one dataset that was not, HarmBench, we deployed two psychometric tests. The responses need more than one dimension, and developer-linked DIF flags are reduced under scope-specific matching, without ruling out genuine domain-specific differences. A number offered as a measure of one attribute should earn that reading before it is used to compare models. For HarmBench, the reading that the psychometric evidence supports is the narrow one that keeps the behaviors apart.

\section{Limitations}
\label{sec:limitations}

Our model pool is a convenience sample of 81 models clustered by developer, not a random draw from a population, and the developer groups are small, with 21 OpenAI and 11 Anthropic models. The DIF result is therefore a statement about these models at this snapshot rather than a general claim about either developer. Models from one developer also share training pipelines and are not independent examinees, so within-developer clustering may inflate the developer DIF we observe under the single score. In addition, our item scores come from two language model judges, so the response matrix inherits whatever biases those judges carry, and a failed item reflects their call rather than ground truth. We did not replicate the analysis using HarmBench's original classifier-based attack-success-rate pipeline for the applicable text items, either for the full dataset or for a subset. Our conclusions therefore apply to HELM's operationalization of HarmBench; the item-level response matrix, dimensional structure, and DIF pattern could differ under HarmBench's original scoring pipeline. We binarize a five-point rubric at a strict threshold, which is conservative but discards gradation that a different cutoff might surface. Finally, our evidence is one benchmark at one release, and while both tests run on any benchmark with item-level responses, we have not yet shown the same structure elsewhere.

\section*{Acknowledgments}
We thank Jeremy N. V. Miles for his invaluable comments on an earlier draft of this manuscript. HELM Safety data are hosted publicly by the Stanford Center for Research on Foundation Models. This work was supported by Google. Any opinions, findings, conclusions, or recommendations expressed in this material are those of the authors and do not reflect the views of Google or other funding agencies.

\bibliography{colm2026_conference}
\bibliographystyle{colm2026_conference}

\appendix
\section{Software, seeds, and reproducibility}
\label{app:repro}

All analyses were performed in Python with the following packages: \texttt{py-irt} for variational MIRT estimation \citep{lalor2023pyirt}, \texttt{pyro} as its inference backend, \texttt{numpy} and \texttt{pandas} for data manipulation, \texttt{scipy} for statistical tests and Procrustes alignment, and \texttt{matplotlib} for figures. Specific versions are listed in the \texttt{requirements.txt} of the repository containing the code for this paper (see below).

For each model and split, we fit twenty random initializations with seeds 0 through 19. Each fit was run for 2{,}000 epochs of stochastic variational inference with the Adam optimizer at learning rate 0.01. Fits that produced NaN losses were automatically retried at learning rate 0.005 with 3{,}000 epochs. Within each model--split cell, we retained the restart with the highest training log-likelihood; held-out responses were never used for restart selection. The predictive entries in Table~\ref{tab:comparison} report the mean and standard deviation across the five retained split-level fits. For the direct comparison between the strongest unidimensional specification and confirmatory 3D, all twenty starts were available in each of the five splits. Across those starts, the mean within-split SD of held-out log-loss was 0.00037 for the unidimensional 3PL and 0.00035 for the confirmatory 3D 2PL. Within every split, every confirmatory 3D restart had lower held-out log-loss than every unidimensional 3PL restart. Fitted parameters for each model, split, and seed combination were cached to disk and used for all these diagnostics.

The notebook and supporting code can be found at \href{https://github.com/cmstewart/harmful-refusal-audit}{https://github.com/cmstewart/harmful-refusal-audit}. The HELM Safety data are publicly hosted by Stanford CRFM and are downloaded by the notebook on first run.

\FloatBarrier

\section{Bernoulli-null eigenvalue check}
\label{app:parallel-analysis}

\begin{figure}[htbp]
\centering
\includegraphics[width=0.62\linewidth]{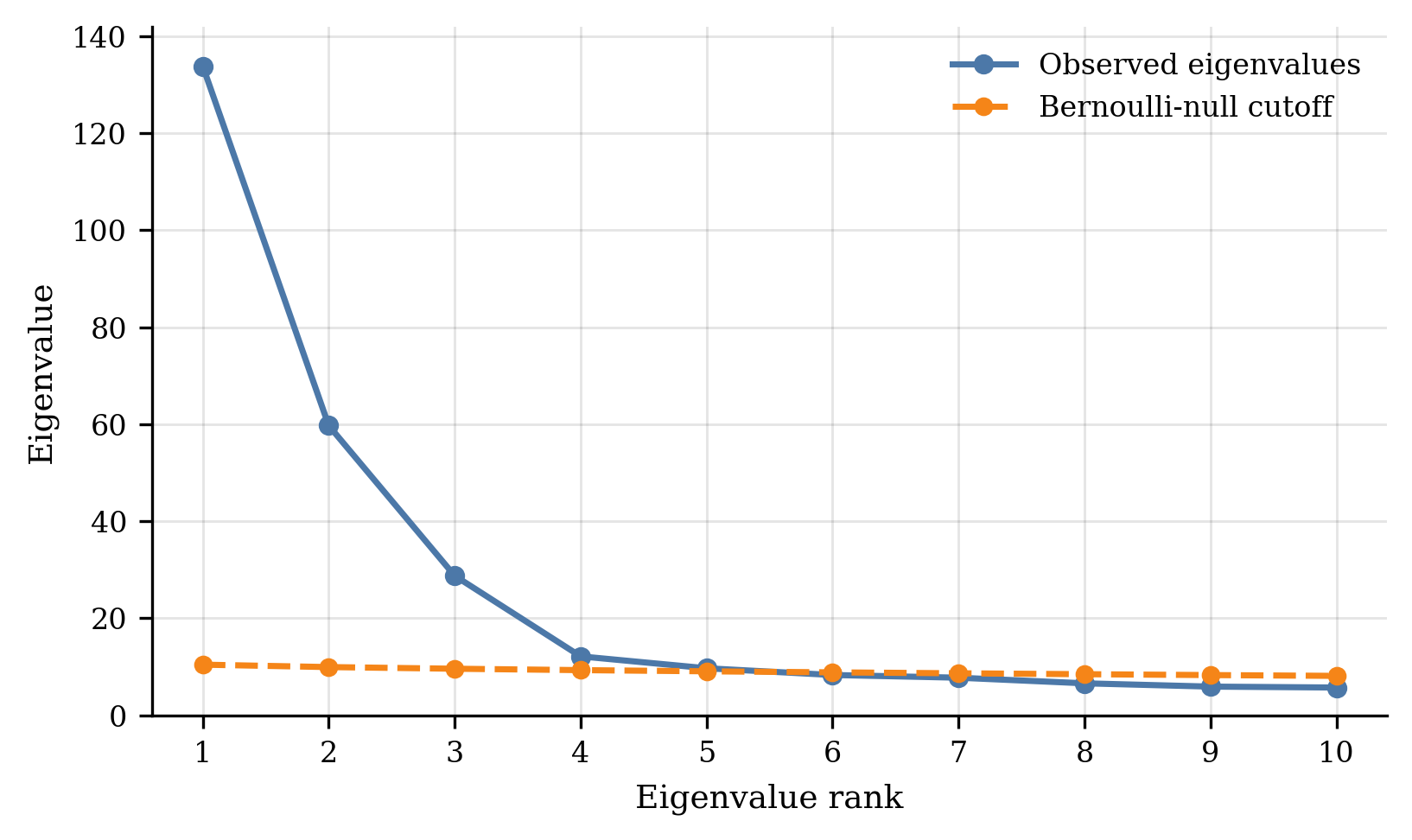}
\caption{Bernoulli-null eigenvalue check of the HarmBench item-correlation matrix. The orange line shows the random-data cutoff from Bernoulli matrices matched to HarmBench's shape and item pass rates. Under one factor plus noise, only the first observed eigenvalue should exceed the cutoff. Several later eigenvalues do as well.}
\label{fig:parallel-analysis}
\end{figure}

Figure~\ref{fig:parallel-analysis} shows the model-light diagnostic summarized in Section~\ref{sec:results}. The plot compares the eigenvalues of the observed HarmBench item-correlation matrix with a random-data cutoff from Bernoulli matrices matched to HarmBench's shape and item pass rates. If one common factor plus noise organized the responses, only the first observed eigenvalue should exceed the cutoff. Instead, several later eigenvalues also exceed it. We use this as a screening result: it rules against a clean one-dimensional interpretation, while the MIRT comparisons in the main text provide the stronger model-based evidence and the interpretable factor structures.

\FloatBarrier

\section{Model-family and content-partition robustness}
\label{app:estimation}

The main analysis treats the confirmatory three-dimensional two-parameter logistic (2PL) response-process model as primary and the confirmatory seven-dimensional HarmBench-domain 2PL model as a sensitivity check. This appendix reports robustness checks that vary two parts of that choice: the item-response family and the item partition. The purpose is not to find the most elaborate model that can improve one predictive metric, but to ask whether the main conclusion changes when the model is made more constrained, more flexible, or more closely aligned with HarmBench's own harm-domain taxonomy.

\subsection{Models compared}

The baseline one-dimensional 2PL model is the strongest version of the single-score interpretation. It assumes that one latent score $\theta_i$ explains model $i$'s responses across all HarmBench items:
\[
P(Y_{ij}=1\mid \theta_i,a_j,b_j)
=
\sigma\{a_j(\theta_i-b_j)\}.
\]
Here $a_j$ is an item-specific discrimination and $b_j$ is an item difficulty/location. The exploratory multidimensional 2PL models in the main comparison generalize this to $d\in\{2,\ldots,10\}$ latent dimensions:
\[
P(Y_{ij}=1\mid \boldsymbol{\theta}_i,\mathbf a_j,\mathbf b_j)
=
\sigma\left\{
\sum_{k=1}^{d}a_{jk}(\theta_{ik}-b_{jk})
\right\}.
\]
These exploratory models let every item load on every dimension. In the unsupervised-learning analogy from the main text, they learn dimensions from the response matrix without preassigned item labels. They are useful for testing whether more dimensions improve prediction, but their dimensions are rotationally ambiguous and, therefore, less directly interpretable.

The preferred confirmatory 3D 2PL model keeps the multidimensional response form but fixes the item-to-factor map in advance. Standard, contextual, and copyright items each load only on their assigned factor. This is a response-process partition: it groups items by what the model is being asked to do, not by HarmBench's harm-domain content labels.

The 7D model is a confirmatory 2PL sensitivity check. It assigns each item to one of HarmBench's seven harm domains: copyright, cybercrime/intrusion, misinformation/disinformation, illegal behavior, chemical/biological harm, harassment/bullying, and general harmful content. This model asks whether HarmBench's own content taxonomy changes the conclusion reached by the more parsimonious three-factor response-process model.

\subsection{Evaluation protocol}

All models use the same HarmBench response matrix, the same five repeated 80/20 response-level train/test splits, and the same seed-selection rule as the main analysis. Held-out log-loss is the primary predictive criterion. Brier score is reported as a secondary calibration-sensitive criterion. AIC and BIC are computed on the training responses and used as in-sample parsimony checks. The parameter count in Table~\ref{tab:model-family-robustness} includes fitted item parameters and fitted model latent scores.

\subsection{Results}

The Rasch check is mainly a floor rather than a serious competitor. A unidimensional 1PL model predicts held-out responses better than the unidimensional 2PL in this run, but it remains far worse than the response-process models. Moving from unidimensional 1PL to confirmatory 3D 1PL reduces held-out log-loss from 0.330 to 0.263. Thus, the need for multiple dimensions is not an artifact of item-specific slopes. Even when all items are forced to discriminate equally, separating standard, contextual, and copyright items substantially improves prediction.

The 3PL check is more informative. A unidimensional 3PL improves over the unidimensional 1PL, but it still does not approach the three-factor models. Adding lower-asymptote parameters to the confirmatory 3D model also does not improve the primary predictive criterion: confirmatory 3D 2PL has held-out log-loss 0.258, compared with 0.261 for confirmatory 3D 3PL. Higher-dimensional exploratory 3PL models slightly improve Brier score, with exploratory 4D 3PL reaching 0.076, but only by adding thousands of fitted quantities and without improving held-out log-loss or the information criteria. The simpler confirmatory 3D 2PL therefore remains preferred over the 3PL variants.

The 7D model is the closest competitor to the preferred model. It slightly improves held-out prediction, with log-loss 0.255 and Brier 0.077, compared with 0.258 and 0.078 for confirmatory 3D 2PL. That small gain must be balanced against the parsimony cost. The confirmatory 3D 2PL has lower AIC and BIC than the 7D content model, 13{,}707 and 22{,}183 versus 13{,}961 and 25{,}080. The 7D result therefore supports the same conclusion in a different way: HarmBench is not one-dimensional, but HarmBench's more granular content partition does not provide enough additional value to displace the simpler standard/contextual/copyright structure.

\begin{table}[htbp]
\centering
\scriptsize
\setlength{\tabcolsep}{4pt}
\renewcommand{\arraystretch}{1.08}
\caption{Model-family and content-partition robustness checks on the same five HarmBench train/test splits used in the main analysis. Params counts fitted item parameters and fitted model latent scores. Lower is better for held-out log-loss, Brier, AIC, and BIC.}
\label{tab:model-family-robustness}
\begin{tabular}{llrrrrrr}
\toprule
Model & Family & $d$ & Params & Log-loss & Brier & AIC & BIC \\
\midrule
Confirmatory 7D & 2PL & 7 & 1{,}363 & \textbf{0.255} & 0.077 & 13{,}961 & 25{,}080 \\
Confirmatory 3D & 2PL & 3 & 1{,}039 & 0.258 & 0.078 & \textbf{13{,}707} & 22{,}183 \\
Exploratory 3D & 3PL & 3 & 3{,}029 & 0.261 & 0.077 & 16{,}722 & 41{,}431 \\
Confirmatory 3D & 3PL & 3 & 1{,}437 & 0.261 & 0.079 & 14{,}726 & 26{,}448 \\
Exploratory 4D & 3PL & 4 & 3{,}906 & 0.262 & \textbf{0.076} & 17{,}873 & 49{,}737 \\
Confirmatory 3D & 1PL/Rasch & 3 & 641 & 0.263 & 0.079 & 13{,}881 & 19{,}110 \\
Unidimensional & 3PL & 1 & 1{,}275 & 0.322 & 0.096 & 17{,}600 & 28{,}001 \\
Unidimensional & 1PL/Rasch & 1 & 479 & 0.330 & 0.100 & 17{,}158 & 21{,}066 \\
Unidimensional & 2PL & 1 & 877 & 0.470 & 0.156 & 25{,}806 & 32{,}961 \\
\bottomrule
\end{tabular}
\end{table}

BIC favors the confirmatory 3D Rasch model over the 3D 2PL because the Rasch model fixes item slopes and is therefore much smaller. We do not treat that as evidence for a single score. The Rasch comparison still requires the three response-process factors to predict well, and the main 2PL analysis remains preferable because it permits HarmBench items to vary in discrimination while retaining a compact, pre-specified factor structure.

Taken together, these checks support the main interpretation. The evidence against a single HarmBench score does not depend on choosing a 2PL model over Rasch, nor does it disappear when the model includes 3PL lower asymptotes. The seven-domain content model confirms that additional structure can squeeze out a small predictive gain, but not enough to justify the added complexity over the more parsimonious three-factor response-process model. The stable result is that one dimension is too weak and that the standard/contextual/copyright structure captures much of the dependence the single score leaves behind.

\FloatBarrier

\section{Differential item functioning details}
\label{app:dif}

The DIF analysis asks whether item responses differ by model group after matching models on the relevant fitted ability. This is a conditional comparison rather than a test of whether two model groups have the same average score. Under the single-score interpretation, two models with the same \emph{harmful refusal} ability should have the same probability of passing a HarmBench item, regardless of which developer built them or whether the model is closed/API-accessed or open-weight-like. Group-linked differences after ability matching are therefore evidence that the item is responding to something other than the target construct.

\subsection{Pre-specified group comparisons}

We use two pre-specified comparisons. The developer comparison is OpenAI versus Anthropic, the only named developers with enough models to support a stable item-level comparison. The second comparison contrasts closed/API models with open-weight-like models. The model counts and raw HarmBench means are shown in Table~\ref{tab:dif-groups}. These raw means are descriptive only. The DIF tests below condition on ability before comparing groups.

\begin{table}[htbp]
\centering
\small
\setlength{\tabcolsep}{8pt}
\renewcommand{\arraystretch}{1.08}
\caption{Pre-specified DIF group comparisons. Means are raw HarmBench pass rates before ability matching.}
\label{tab:dif-groups}
\begin{tabular}{llrr}
\toprule
Comparison & Group & Models & Mean score \\
\midrule
Developer & OpenAI & 21 & 0.866 \\
Developer & Anthropic & 11 & 0.889 \\
Access type & Closed/API & 48 & 0.776 \\
Access type & Open-weight-like & 33 & 0.507 \\
\bottomrule
\end{tabular}
\end{table}

\subsection{Mantel-Haenszel screen}

The primary DIF screen uses the Mantel-Haenszel (MH) common odds ratio. The practical idea is to compare like with like. We first put models into coarse ability bands, then ask within each band whether one group passes a particular item more often than the other group. If the groups have similar pass odds inside the bands, the item behaves like the same biased coin for both groups at that ability level. If the odds remain different across the bands, the item is calibrated differently for the groups.

Within each scope, models are sorted into three ability bands. The single-score scope uses the unidimensional HarmBench ability estimate. The three response-process scopes use the corresponding fitted factor score for standard, contextual, or copyright items, and the seven-domain sensitivity uses the corresponding fitted HarmBench-domain score. Within ability band $s$ and item $j$, let $A_{js}$ and $B_{js}$ be the pass and fail counts for group A, and let $C_{js}$ and $D_{js}$ be the pass and fail counts for group B. The common odds ratio is
\[
\widehat{\alpha}_{MH,j}
=
\frac{\sum_s A_{js}D_{js}/N_{js}}
{\sum_s B_{js}C_{js}/N_{js}},
\]
where $N_{js}=A_{js}+B_{js}+C_{js}+D_{js}$. We use $|\log \widehat{\alpha}_{MH,j}|$ as the item statistic.

The cutoff is calibrated by parametric bootstrap rather than by a nonparametric resample. For each scope and comparison, we simulate 5{,}000 no-DIF response matrices from the fitted IRT model while keeping the observed group labels fixed. We then rerun the same MH pipeline on every simulated matrix. This yields item-specific null cutoffs and a family-wise error rate (FWER) cutoff. The FWER cutoff is the 95th percentile of the maximum item statistic observed anywhere in the same testing family under the no-DIF bootstrap, so it is stricter than an item-specific cutoff.

\subsection{Logistic sensitivity screen}

As a sensitivity check, we also fit an item-level ridge-logistic DIF model. This asks the same practical question without cutting ability into bands. For each item, the model estimates how pass probability changes with fitted ability and then asks whether the group label still shifts that probability after ability is included. A large group coefficient means that two models at the same fitted ability would still get different predicted probabilities for the same item.

The model is
\[
\operatorname{logit}\{P(Y_{ij}=1)\}
=
\alpha_j+\beta_j \widehat{\theta}_{i}+\gamma_j G_i,
\]
where $\widehat{\theta}_{i}$ is the relevant fitted ability and $G_i$ is the group indicator. The DIF statistic is $|\gamma_j|$. This approach keeps ability continuous rather than splitting it into bands. The ridge penalty stabilizes estimates in this small model sample, especially for near-separated items. The logistic screen uses the same no-DIF parametric-bootstrap logic as the MH screen and is treated as corroborating evidence rather than as a replacement for MH.

\subsection{Main DIF results}

Table~\ref{tab:dif-mh-logistic} summarizes the primary MH screen beside the logistic sensitivity check. The signal is concentrated in the single-score OpenAI-versus-Anthropic comparison. Under the unidimensional HarmBench score, MH flags 13 items at the FWER q95 level and the logistic screen flags 17. Once items are scored within the three response-process scopes, strict FWER-level developer DIF mostly disappears: MH leaves one contextual flag, and logistic leaves two standard-item flags. The access-type comparison has no FWER-level flags in any single-score or response-process scope, even though the two access-type groups differ substantially in raw mean score.

\begin{table}[htbp]
\centering
\scriptsize
\setlength{\tabcolsep}{4pt}
\renewcommand{\arraystretch}{1.08}
\caption{DIF flags under Mantel-Haenszel and ridge-logistic parametric bootstraps. Item q95 counts use item-specific 95th-percentile null cutoffs. FWER q95 counts use the stricter max-statistic family-wise cutoff.}
\label{tab:dif-mh-logistic}
\begin{tabular}{llrrrrr}
\toprule
Scope & Comparison & Items & MH item q95 & MH FWER & Logit item q95 & Logit FWER \\
\midrule
Single score & OpenAI vs.\ Anthropic & 398 & 94 & 13 & 102 & 17 \\
Single score & Closed/API vs.\ open & 398 & 41 & 0 & 51 & 0 \\
Standard & OpenAI vs.\ Anthropic & 199 & 27 & 0 & 25 & 2 \\
Standard & Closed/API vs.\ open & 199 & 13 & 0 & 12 & 0 \\
Contextual & OpenAI vs.\ Anthropic & 99 & 16 & 1 & 15 & 0 \\
Contextual & Closed/API vs.\ open & 99 & 8 & 0 & 8 & 0 \\
Copyright & OpenAI vs.\ Anthropic & 100 & 7 & 0 & 6 & 0 \\
Copyright & Closed/API vs.\ open & 100 & 16 & 0 & 15 & 0 \\
\bottomrule
\end{tabular}
\end{table}

The contrast between item-specific and FWER counts is useful diagnostically. Item-specific q95 flags show that group-linked texture remains in several scopes, especially for the access-type comparison in copyright items. But the FWER counts show that this texture is not strong enough to support a broad claim of DIF once the response-process scopes are separated. The main construct-validity result is therefore the collapse of broad developer DIF when HarmBench is not forced into one undifferentiated score.

\subsection{Seven-domain sensitivity}

We also reran the DIF screens within the HarmBench seven-domain confirmatory scopes. This sensitivity check asks whether the response-process result is an artifact of ignoring HarmBench's own content labels. Table~\ref{tab:dif-domain7} reports FWER q95 flags only. Most domains show no FWER-level DIF after ability matching. The notable exception is cybercrime/intrusion, where OpenAI-versus-Anthropic flags remain under both MH and logistic screens. The logistic screen also leaves one OpenAI-versus-Anthropic flag in misinformation/disinformation. No seven-domain access-type comparison has an FWER-level flag.

\begin{table}[htbp]
\centering
\scriptsize
\setlength{\tabcolsep}{5pt}
\renewcommand{\arraystretch}{1.08}
\caption{Seven-domain DIF sensitivity. Entries are FWER q95 flag counts. The developer comparison is OpenAI versus Anthropic; the access-type comparison is closed/API versus open-weight-like.}
\label{tab:dif-domain7}
\begin{tabular}{lrrrrr}
\toprule
Domain & Items & Dev. MH & Dev. logit & Access MH & Access logit \\
\midrule
Chemical/biological & 56 & 0 & 0 & 0 & 0 \\
Copyright & 100 & 0 & 0 & 0 & 0 \\
Cybercrime/intrusion & 67 & 2 & 4 & 0 & 0 \\
Harassment/bullying & 25 & 0 & 0 & 0 & 0 \\
General harmful content & 22 & 0 & 0 & 0 & 0 \\
Illegal behavior & 63 & 0 & 0 & 0 & 0 \\
Misinformation/disinformation & 65 & 0 & 1 & 0 & 0 \\
\bottomrule
\end{tabular}
\end{table}

The seven-domain sensitivity does not change the main conclusion. DIF is strongest when HarmBench is treated as one undifferentiated scale. Most FWER-level DIF is absorbed once items are scored within response-process or content-domain scopes, with cybercrime/intrusion remaining as a content-domain sensitivity finding for the OpenAI-versus-Anthropic comparison.

\section{Model list and cohort composition}
\label{app:models}

Table~\ref{tab:models} lists the 81 models in the analysis. The leaderboard listed 87 models. We excluded six for incomplete per-item data on at least one benchmark, leaving 81.

\par\vspace{1em}
\begingroup
\footnotesize
\captionof{table}{All 81 models in the analysis (HELM Safety v1.17.0).}
\label{tab:models}
\begin{multicols}{2}
\raggedright
\ttfamily
anthropic\_claude-3-5-sonnet-20240620 \\
anthropic\_claude-3-haiku-20240307 \\
anthropic\_claude-3-opus-20240229 \\
anthropic\_claude-3-sonnet-20240229 \\
cohere\_command-r \\
cohere\_command-r-plus \\
databricks\_dbrx-instruct \\
deepseek-ai\_deepseek-llm-67b-chat \\
google\_gemini-1.5-flash-001 \\
google\_gemini-1.5-pro-001 \\
meta\_llama-3-70b-chat \\
meta\_llama-3-8b-chat \\
meta\_llama-3.1-405b-instruct-turbo \\
meta\_llama-3.1-70b-instruct-turbo \\
meta\_llama-3.1-8b-instruct-turbo \\
mistralai\_mistral-7b-instruct-v0.1 \\
mistralai\_mistral-7b-instruct-v0.3 \\
mistralai\_mixtral-8x22b-instruct-v0.1 \\
mistralai\_mixtral-8x7b-instruct-v0.1 \\
openai\_gpt-3.5-turbo-0125 \\
openai\_gpt-3.5-turbo-0613 \\
openai\_gpt-3.5-turbo-1106 \\
openai\_gpt-4-turbo-2024-04-09 \\
openai\_gpt-4o-2024-05-13 \\
openai\_gpt-4o-mini-2024-07-18 \\
qwen\_qwen1.5-72b-chat \\
qwen\_qwen2-72b-instruct \\
deepseek-ai\_deepseek-r1 \\
deepseek-ai\_deepseek-r1-hide-reasoning \\
deepseek-ai\_deepseek-v3 \\
openai\_o1-2024-12-17 \\
openai\_o1-mini-2024-09-12 \\
openai\_o3-mini-2025-01-31 \\
anthropic\_claude-3-7-sonnet-20250219 \\
openai\_gpt-4.5-preview-2025-02-27 \\
writer\_palmyra-fin \\
writer\_palmyra-x-004 \\
meta\_llama-4-maverick-17b-128e-instruct-fp8 \\
meta\_llama-4-scout-17b-16e-instruct \\
openai\_gpt-4.1-2025-04-14 \\
openai\_gpt-4.1-mini-2025-04-14 \\
openai\_gpt-4.1-nano-2025-04-14 \\
xai\_grok-3-beta \\
xai\_grok-3-mini-beta \\
google\_gemini-2.5-flash-preview-04-17 \\
google\_gemini-2.5-pro-preview-03-25 \\
openai\_o3-2025-04-16 \\
openai\_o4-mini-2025-04-16 \\
qwen\_qwen3-235b-a22b-fp8-tput \\
writer\_palmyra-med \\
writer\_palmyra-x5 \\
anthropic\_claude-opus-4-20250514 \\
anthropic\_claude-opus-4-20250514-thinking-10k \\
anthropic\_claude-sonnet-4-20250514 \\
anthropic\_claude-sonnet-4-20250514-thinking-10k \\
deepseek-ai\_deepseek-r1-0528 \\
allenai\_olmo-2-0325-32b-instruct \\
allenai\_olmo-2-1124-13b-instruct \\
allenai\_olmo-2-1124-7b-instruct \\
allenai\_olmoe-1b-7b-0125-instruct \\
marin-community\_marin-8b-instruct \\
moonshotai\_kimi-k2-instruct \\
xai\_grok-4-0709 \\
ibm\_granite-3.3-8b-instruct \\
google\_gemini-2.5-flash-lite \\
openai\_gpt-oss-120b \\
zai-org\_glm-4.5-air-fp8 \\
openai\_gpt-5-2025-08-07 \\
openai\_gpt-5-mini-2025-08-07 \\
openai\_gpt-5-nano-2025-08-07 \\
openai\_gpt-oss-20b \\
qwen\_qwen3-235b-a22b-instruct-2507-fp8 \\
anthropic\_claude-sonnet-4-5-20250929 \\
qwen\_qwen3-next-80b-a3b-thinking \\
ibm\_granite-4.0-h-small \\
ibm\_granite-4.0-h-small-with-guardian \\
ibm\_granite-4.0-micro \\
ibm\_granite-4.0-micro-with-guardian \\
anthropic\_claude-haiku-4-5-20251001 \\
google\_gemini-3-pro-preview \\
openai\_gpt-5.1-2025-11-13 \\
\end{multicols}
\endgroup

\end{document}